\documentclass[conference]{IEEEtran}
\IEEEoverridecommandlockouts
\usepackage{cite}
\usepackage{amsmath,amssymb,amsfonts}
\usepackage{algorithmic}
\usepackage{graphicx}
\usepackage{textcomp}
\usepackage{xcolor}
\usepackage[norelsize]{algorithm2e}
\usepackage[a4paper, total={7in, 10in}]{geometry}
\def\BibTeX{{\rm B\kern-.05em{\sc i\kern-.025em b}\kern-.08em
    T\kern-.1667em\lower.7ex\hbox{E}\kern-.125emX}}
\begin{document}

\title{A Multi-language Platform for Generating Algebraic Mathematical Word Problems}

\author{\IEEEauthorblockN{ Vijini Liyanage}
\IEEEauthorblockA{\textit{Department of Computer Science \& Engineering } \\
\textit{University of Moratuwa}\\
Katubedda 10400, Sri Lanka \\
vijiniliyanage.12@cse.mrt.ac.lk}

\and
\IEEEauthorblockN{ Surangika Ranathunga}
\IEEEauthorblockA{\textit{Department of Computer Science \& Engineering } \\
\textit{University of Moratuwa}\\
Katubedda 10400, Sri Lanka \\
surangika@cse.mrt.ac.lk}
}
\IEEEoverridecommandlockouts

\IEEEpubid{\makebox[\columnwidth]{978-1-7281-3706-3/19/\$31.00~\copyright{}2019 IEEE \hfill} \hspace{\columnsep}\makebox[\columnwidth]{ }}

\maketitle
\IEEEpubidadjcol
\begin{abstract}
Existing approaches for automatically generating mathematical word problems are deprived of customizability and creativity due to the inherent nature of template-based mechanisms they employ. We present a solution to this problem with the use of deep neural language generation mechanisms. Our approach uses a Character Level Long Short Term Memory Network (LSTM) to generate word problems, and uses POS (Part of Speech) tags to resolve the constraints found in the generated problems. Our approach is capable of generating Mathematics Word Problems in both English and Sinhala languages with an accuracy over 90\%.

\end{abstract}


\section{Introduction}\label{Section I}
A Mathematical word problem (MWP) is a mathematical problem expressed in natural language. Unlike other knowledge based question types such as travel or history related questions, MWPs require problem solving ability. In particular, algebraic questions involve sentences to make the questions more deep and inspective. Algebra is a major component of mathematics that is learnt by every student in Ordinary Level (O/L). Simple algebra problems mostly appear in a word format. \lq Kamal has 16 marbles and Nimal has 12 less marbles than Kamal, how many marbles does Nimal have \rq is an example for a simple algebra problem.

However, a majority of students face difficulties in solving MWPs related to algebra [1],[2]. The most effective way to mitigate this problem is to provide the students with a lot of similar problems to work on. However generating a large number of fresh word problems is a tedious and time consuming task for the teachers. According to researchers, the integration of Information Communication Technology (ICT) in algebra education suggests a positive influence on student achievements in general [3]. Therefore developing a system that can automatically generate algebra problems is a timely requirement.

Automatic mathematics question generation has recently drawn the efforts of researchers in a number of arenas including algebra, geometry and statistics [4-9]. However, the existing approaches for algebra word problem  generation are fully or semi template based [4-9], which makes the questions formed look alike. Therefore prevailing systems restrict the creativity, novelty and the multilingualism of the generated questions. Although Natural Language Generation (NLG) with Generative models is quite popular in the modern research arena [10-14], we are not aware of any existing systems that use generative model for Mathematics question generation.

We present a solution to this problem with the use of deep neural language generation mechanisms. Most of the Algebraic questions contain numerical constraints. For example in the problem, `Harry has 9 oranges and Mary has 3 less oranges than Harry, how many oranges does Mary have'', the first numerical value should be higher than the second numerical value. Our system is capable of handling this kind of constraints.

A character level Long Short Term Memory Network (LSTM) was used for the word problem generation. Generated questions were filtered considering Part of Speech (POS) to satisfy the numerical constraints. A dataset with 1350 English language MWPs for the Elementary Level, a dataset with 2350 English language MWPs for GCE Ordinary Level, and another dataset with 500 Sinhala language MWPs for Elementary Level  were created. These datasets were used to train the model. Our system is capable of generating multilingual problems because depending on the language of the dataset, the system can generate problems in that particular language.\\
Questions generated by the current system are not 100\% accurate. Therefore we conducted a verification test with a group of tutors to prove that our system is more efficient than manually producing fresh problems by the tutors. The tutors were asked to modify the generated questions in order to make them 100\% accurate. They were also asked to produce new questions from scratch. It was proved that the process of generating problems by our system and applying minor modifications is still efficient that manually producing the problems. Our system achieved an accuracy over 90\% for the problems generated in English and Sinhala languages.

The rest of the paper is organized as follows. Related work is discussed in section II. The identified neural generative models are described in section III. The characteristics and constraints of the datasets are described in section IV. Methodology is  described in section V.  Evaluation results of the system are provided in section VI and Conclusion and future work are mentioned in section VII.

\section{Related Work}\label{Section II}

In the domain of math word problem generation, among the existing solutions, the most common automation approach is  template-based database approach. This approach has proved itself valuable for word problem generation, but the personalization level is insufficient for engaging education since the generated problems tend to follow a similar pattern. Deane and Sheehan [4] built such an automatic word problem generating mechanism, where natural language generation happens with Frame Semantics. Polozov\rq s [5] approach also has been built upon the same architecture for NLG, but in addition the word problem generating logic includes Answer Set Programming to satisfy a collection of pedagogical and narrative requirements. The mentioned two research depends on templates to finally produce the problem. Singh et al [6] proposed a semi automated template based approach for algebra proof problems. But the mentioned theme oriented approaches are restricted to generate problems only with the chosen templates, which will restrict the generated problems to follow a certain pattern and eliminate the creativity.

In order to motivate students to engage in word problems, a theme rewriting approach was proposed by Koncel-Kedziorski et al [7], which rewrites the same question in more interesting themes such as Star Wars. But this approach does not generate fresh problems. Relatively few research studies addressed the problem in generating fresh mathematical word problems. For example, Williams et al [8] used the Web Ontology Language to represent such problems. English statements are then extracted from this knowledge representation. However, this approach produces limited types of word problems, in which the difficulty level is controlled merely by changing the generated sentence or by adding some distraction to the sentence.

Wang et al.[9], have leveraged the concept of expression trees to generate math word problems. The tree structure can provide the skeleton of the story, and meanwhile allows the story to be constructed recursively from the sub-stories. Each sub-story can be seen as a text template with value slots to be filled. These sub-stories are concatenated into an entire narrative. Although the proposed solution proves to be a step forward than the other existing systems due to its capability in generating authentic, diverse and configurable mathematical word problems, this approach strictly depends on the dimensional units and templates to generate the expression trees and to derive sub stories for the generated Atomic Expression Trees, respectively.

We are not aware of any existing approach that uses state-of-the-art neural generation models (discussed in the section III) to generate Mathematics word problems.

\section{Neural Generative Models}\label{Section III}
According to the latest research [10, 15], text generation has been performed using either of the three models; 1. Auto-regressive or maximum likelihood estimation (MLE)-based models [11], 2. Reinforcement learning (RL)-based approaches [16] and 3. Generative Adversarial Networks (GAN) [17].

Auto-regressive models refer to Recurrent Neural Networks (RNNs), at its most fundamental level. It is simply a type of densely connected neural network. The key difference to normal feed forward networks is the introduction of time -–  the output of the hidden layer in an RNN is fed back into itself. For RNNs, ideally, long memories are preferred, so the network can connect data relationships at significant distances in time. The more time steps the RNN has, the more chance RNN has back-propagation gradients either accumulating and exploding or vanishing down to nothing, introducing  the exploding/vanishing (respectively) gradient problem. This issue has been answered through the introduction of LSTM (Long Short Term Memory Networks).

The LSTM cell reduces the vanishing gradient problem by creating an internal memory state and adding it to the processed input, which greatly reduces the multiplicative effect of small gradients. The time dependence and effects of previous inputs are controlled by an interesting concept called a forget gate, which determines which states are remembered or forgotten. When LSTMs are used for text generation, it learns the likelihood of occurrence of a word/character based on the previous sequence of words/characters used in the text. Text generation process with a Character Level LSTM is elaborated in section IV.

Merity et al.[27] analyzed two types of LSTMs, Character Level LSTM and Word Level LSTM, which generate the next character and the next word of the sequence respectively. Since Word Level LSTMs suffer from increased computational cost due to large vocabulary sizes and that Word Level LSTMs need to replace infrequent words with Out-Of-Vocabulary(OOV) tokens, they have stated that Character Level LSTMs perform better than Word Level LSTMs. But on the other hand Character Level LSTMs are slower to process than Word Level LSTMs, as the number of tokens increases substantially. But this research[27] shows that an adaptive softmax is capable of modeling both character level and word level LSTMs enabling to achieve state-of-the-art results.

Merity et.al\rq s [27] research has optimized the word level LSTM model using some techniques such as the use of DropConnect[28] on the recurrent hidden to hidden weight matrices, and the use of Average Stochastic Gradient Descent (ASGD) to further improve the training process. Here, the first technique prevents overfitting on the recurrent connections of the LSTM. When training with Dropout (a regularization mechanism used by many previous work), a randomly selected subset of activations are set to zero within each layer. DropConnect instead sets a randomly selected subset of weights within the network to zero. The advantage here is that DropConnect does not require any modifications to an RNN's formulation. As the dropout operation is applied once to the weight matrices, before the forward and backward pass, the impact on training speed is minimal and any standard RNN implementation can be used.

In RL (Reinforcement Learning), the goal of the agent is to examine the state and the reward information it receives, and choose an action that maximizes the reward feedback it receives. RL is a gradual stamping of behavior that comes from receiving rewards and punishments (negative rewards). When using RL for text generation, the actions are writing words and the states are the words the algorithm has already written. Choosing the best word to write is hard because there are as many actions as there are words in the vocabulary. A kind of reinforcement learning that works well in domains with large action spaces such as text generation is called policy gradient[18]. A policy specifies what action to take (what word to write) for each state. One difficulty with these high dimensional spaces is that it is hard to learn what action to take (what word to write) when the model doesn\rq t get a reward.

A GAN (Generative Adversarial Network) consists of two neural networks, the generator that generates new data instances and the discriminator that evaluates them for authenticity. I.e. the discriminator decides whether each instance of data that it reviews belongs to the actual training dataset or not. Both nets are trying to optimize a different and opposing objective function, or loss function. GANs can undergo mode collapse issue, which means once the generator figures out how to fool the discriminator, it may keep generating the same thing over and over again. GANs are also hard to train because it is difficult to keep the generator and discriminator in balance. There are mainly two types of GANs for text generation [19],
 \begin{itemize}
\item GANs that use reinforcement learning (RL) algorithms for text generation such as SeqGAN[20], MaliGAN[21], RankGAN[22], MaskGAN[23] and LeakGAN[24]. But these are slow to train due to their complex design.

\item GANs in the RL-free category such as GSGAN[25] and TextGAN[26], which use the Gumbel-softmax and soft-argmax trick, respectively, to deal with discrete data. They may suffer from gradient-vanishing issue of the discriminator as a result of keeping the original GAN loss function.
  \end{itemize}

\section{Dataset}\label{Section IV}

We created three datasets of single sentence MWPs, which include a dataset consisting of 1350 elementary level English Medium MWPs, a dataset consisting of 2350 GCE Ordinary Level English Medium MWPs and a dataset consisting 500 elementary level Sinhala Medium MWPs. The two elementary level datasets contain questions with several constraints such as,
\begin{enumerate}
  \item  The first numerical value should be higher than the second numerical value.

Eg: Harry has 9 oranges and Mary has 3 less oranges than Harry, how many oranges does Mary have

  \item When there are questions with units, the questions should use the units that are relevant for the respective items.

Eg: Dina made cookies and she used 0.625 kg flour and 0.25 kg sugar, how much more flour than sugar did Dina use

\item As shown in the previous example, the quantities combined in the question should match with each other.

Eg: flour and sugar for cake.

\item In some problems, the values should not invalidate mathematical concepts.

 E.g.: Three consecutive integers have the sum of 153, what is the second integer -
 Here the numerical value should be chosen such that all the three consecutive numbers remain as integers.

\end{enumerate}

\section{Methodology}\label{Section V}
Recent research[29, 30]  has shown that a well-adjusted MLE (Maximum Likelihood ) model outperforms the considered GAN architectures. Temperature tuning has facilitated the models using MLE (Maximum Likelihood) to outperform the GANs. On the other hand, when using reinforcement learning techniques to generate text, it is hard to learn what action to take (what word to write) when the model does not get a reward. When the dataset is comparatively small, the GAN models tend to produce meaningless outputs. 

Despite the above observations, we experimented with both GAN and MLE models. We used TextGAN [26] to generate the MWPs, which incorporates Gumbel-softmax and soft-argmax trick, instead of RL based GANS that are slower to train [19].\\
Under the MLE approach, we used both word level and character level LSTMs to generate word problems. As shown in the next section, the character level LSTM outperformed the word level LSTM. It also consumed a relatively low time span. Therefore we chose Character Level LSTM over the other models, to be considered for further improvements. We have used Part of Speech (POS) tags as a post processing technique to improve the accuracy of the text generated by the character level LSTM. Especially when there are constraints, the POS tag approach finds the instances to be altered. 

\begin{figure}[htbp]
\centerline{\includegraphics[width=75mm,scale=1.5]{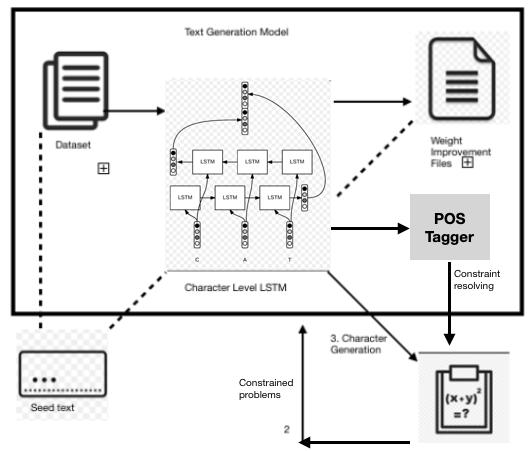}}
\caption{The system diagram.}
\label{fig1}
\end{figure}

As depicted in Fig 1, we have trained the Character Level LSTM model with each of the three datasets, and their trained models are saved separately. Per each of the datasets, the input to output pairs are encoded as integers. The identified patterns are then reshaped, normalized and one hot encoded to the output variables. The LSTM model was defined as a sequential model and the Dropout regularization mechanism is used to randomly select activations and set them to zero and DropConnect was used to randomly select weight matrices between hidden layers and to set them to zero. Then the model was trained with 15-20 epochs depending on the weight improvement.

The generation process starts by randomly selecting a seed text from the dataset. The length of the seed can be 20 - 30 characters, depending on the sequence size of the patterns we have selected previously. From the last character in the seed sequence, the model will be generating the rest of the characters and build the MWP. The model is capable of generating more than one problem at a time, depending on the range of characters defined in the code.

During the text generation process, a naive approach applies greedy sampling, which always chooses the most likely character from the softmax output of the model. But such an approach kills the creativity and novelty of the sequences generated thereby producing repetitive, predictable sequences. Therefore it is important to introduce randomness in the sampling process of the probability distribution for the next character. This process is known as Stochastic Sampling[31]. On the other hand, too much randomness or entropy will produce characters that will add no meaning to the sequence. Therefore in order to control the randomness of the generation process, our approach uses a parameter called the softmax temperature[34]. This temperature parameter is able to characterize the entropy of the probability distribution used for sampling. Given a temperature value, a new probability distribution is computed from the original distribution by re-weighting it. Therefore by introducing a temperature parameter, our approach is capable of generating creative yet realistic sequences of text.

Generation of problems consisting different patterns from one another is enabled by randomizing the selection of seed text and incorporating temperature tuning. When unique problems are required, the entropy of the system is increased by adjusting the temperature tuning parameter. The dropout regularization used in the LSTM also contributes in randomizing the output by reducing the over-fitting problem. Through these mechanisms, our approach is capable of generating creative and innovative word problems, in comparison to the existing template based approaches.

As mentioned in section V, there are some questions that contain numerical constraints. For example, consider the question \lq Harry has 9 oranges and Mary has 3 less oranges than Harry, how many oranges does Mary have?\rq. Here the first numerical value should be greater than the second numerical value.
Some of the generated questions as the one given below had violated the numerical constraint requirements.

  vimal built house and he used 2 kg cement and 6 kg water, how much more cement than water did vimal use

In the above question, there is a problem with the quantities. Since the question states \lq more cement than water\rq, the numeric value that represents cement quantity should be higher than that for water.\\
In Senrich et.al\rq s research [32], it has been proved that the input features such as POS tags can improve the accuracy of the text sequences that are generated by neural networks. \\
Therefore in order to identify and solve such problems our approach uses Natural Language Toolkit (NLTK) POS filtering mechanism for the questions generated in English language. The post processing POS mechanism shown in Algorithm 1 is used to identify the numeric values, units and adjectives such as \lq more\rq or \lq less\rq, which are followed by the preposition \lq than\rq. Then our algorithm compares the numeric values with each other, in relation to the adjective and preposition combination. If any contradiction is found, those will be resolved using the algorithm. Another problem found in this example is that the units used are inappropriate (kg is used to represent amount of water). In order to resolve such issues, our algorithm focuses on consecutive noun pairs that come after the number tags, and check whether they match with each other. Since the accuracy of the existing taggers are not good [35],[36], we did not use POS tag filtering to resolve constraints found in generated Sinhala problems.

\begin{algorithm}
  \SetAlgoLined
  \KwData{Generated question}
  \KwResult{Constrained satisfied question }
  initialization\;
  $tokens$ = $word\_tokenize(Data)$

  $nltk.pos\_tag(tokens)$

  \If{$POS\_tag\_sequence.contains(2\*CD AND JJR)$}{
    \If{$JJR$ = \lq more\rq}{
        \While{$first\_CD\_value$ $\gneq$ $second\_CD\_value$}{
        $first\_CD\_value++$
        }
        \eIf{$units\_dictionary.contains(NN\_Bigrams)$}{
        Output(Data)\;
        }
        {
        modify(NN\_Bigrams)\;
        Output(Data)\;
        }
      }
    }
    \label{algorithm 1}
    \caption{The Algorithm to solve constraints identified in problems}
\end{algorithm}

After applying the Algorithm 1 to solve the constraints, the constrained issues were 100\% removed.
An example for a generated question before applying constraint satisfaction:
 \begin{itemize} \label{question1}
 \item
vimal built house and he used 2 kg cement and 6 kg water, how much more cement than water did vimal use$?$
\end{itemize}
   The corrected problem after applying constraint satisfaction,
   \begin{itemize} \label{question2}
   \item
vimal built house and he used 7 kg cement and 6 l water, how much more cement than water did vimal use$?$
\end{itemize}

\section{Results and Evaluation}\label{Section VI}
The BLEU score results are depicted in TABLE 1. BLEU [33] is an automated, language-independent and fast evaluation metric. It compares modified n-grams of the candidate (generated text) with the modified n-grams of the reference dataset and count the number of matches.

TextGAN took several days to train the model even with a small dataset, yet yielding only a BLEU-2 (Bi-Lingual Evaluation Understudy) score of 0.012. Compared to GAN, there was a huge improvement in the accuracy of the generated text with the use of LSTM (A minimum BLEU-2 score of 0.23). The Character Level LSTM outperformed the other considered models in terms of BLEU score. The average BLEU scores for word level and character level LSTMs regarding the generation of English MWPs were 0.1325 and 0.5025, respectively. The BLEU scores of the generated MWPs with different models in English and Sinhala languages are depicted in Table 1 and Table 2, respectively.

\begin{table}[htbp]
\caption{Bleu Scores Produced By Different Models regarding the formation of English MWPs }
\begin{center}
\begin{tabular}{|c|c|c|c|c|}
\hline
\textbf{Model}&\textbf{BLEU 2}&\textbf{BLEU 3}&\textbf{BLEU 4}&\textbf{BLEU 5} \\

\hline
TextGAN & 0.012 & 0.00 & 0.00 & 0.00  \\
\hline

Simple Word \\Level LSTM & 0.23 & 0.15 & 0.08 & 0.00  \\
\hline

Simple Character\\ Level LSTM & 0.37 & 0.24 & 0.19 & 0.17  \\
\hline

Optimized Word\\ Level LSTM \\(Merity et al.,2018) & 0.31 & 0.15 & 0.07 & 0.00  \\
\hline

Optimized Character\\ Level LSTM \\(Our approach) & 0.69 & 0.58 & 0.42 & 0.32  \\
\hline
Optimized Character\\ Level LSTM \\ after applying\\ POS based\\ post processing & 1.0 & 0.98 & 0.87 & 0.80  \\
\hline

\end{tabular}
\label{tab1}
\end{center}
\end{table}

\begin{table}[htbp]
\caption{Bleu Scores Produced By Different Models regarding the formation of Sinhala MWPs}
\begin{center}
\begin{tabular}{|l|l|l|l|l|}
  \hline
  Model & BLEU 2 & BLEU 3 & BLEU 4 & BLEU 5\\

  \hline
  Simple Word \\Level LSTM & 0.09 & 0.07 & 0.00 & 0.00  \\
  \hline
  Simple Character\\ Level LSTM & 0.22 & 0.19 & 0.09 & 0.04  \\
  \hline
  Optimized Word\\ Level LSTM \\(Merity et al.,2018) & 0.27 & 0.09 & 0.00 & 0.00  \\
  \hline
  Optimized Character\\ Level LSTM \\(Our approach) & 0.30 & 0.19 & 0.17 & 0.09  \\
  \hline

\end{tabular}
\label{tab2}
\end{center}
\end{table}

As the generated questions were not 100\% well-formed, the usability and effectiveness of the system was evaluated using a group of 4 Mathematics tutors. The tutors were asked to generate single sentence mathematics word problems in both English and Sinhala languages. Each tutor created a sample of ten word problems in each language and the respective times for generations were calculated. Then the same set of tutors were asked to correct any mistake in the word problems generated by the system. Each tutor corrected ten word problems in each language that were generated by the system and the respective times were calculated. According to the results provided by tutors, the system is capable of delivering accurate word problems to the students more than 80\% faster than those word problems been produced manually by the tutors. The results provided by each tutor is depicted in TABLE 3.

\begin{table}[htbp]
\caption{Evaluations Done By The Tutors}
\begin{center}
\begin{tabular}{|l|l|l|l|l|}
  \hline
    & Time & Time & Time & Time\\
    & to & to & to & to\\
    & generate & correct & generate & correct\\
    & 10 & 10 & 10 & 10\\
    & English & English & Sinhala & Sinhala\\
    & MWPs & MWPs & MWPs & MWPs\\
    & (minutes) & (minutes) & (minutes) & (minutes)\\
  \hline
  Tutor1 & 18 & 3 & 25 & 2 \\
  \hline
  Tutor2 & 20 & 2.5 & 22 & 4 \\
  \hline
  Tutor3 & 15 & 2 & 15 & 3 \\
  \hline
  Tutor4 & 15 & 3 & 17 & 2.5 \\
  \hline
  Tutor5 & 21 & 4 & 20 & 5 \\
  \hline
  Average & 17.8 & 2.9 & 19.8 & 3.9 \\
  \hline
\end{tabular}
\label{tab3}
\end{center}
\end{table}

The application of POS filtering as a post processing technique made it possible to generate problems which were 100\% accurate. Sample sets of English problems before and after applying the POS algorithm are provided in Fig 3 and Fig 4 respectively. The constraint with amounts $(Eg 2 and 6)$ was resolved $(now 5 and 3)$ and the constraint with units (kg is not suitable for water) was resolved by changing the unit to l. Therefore post processing using POS tags could eliminate the problems related to numerical constraints as well as problems related to units.

\begin{figure}[htbp]

\centerline{\includegraphics[width=75mm,scale=1.5]{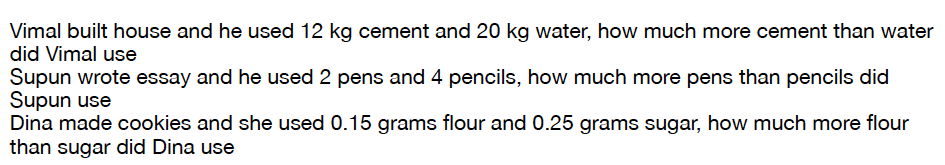}}
\caption{A sample set of English questions generated by the system (Before applying constraint satisfaction with POS tags)}
\label{fig3}
\end{figure}

\begin{figure}[htbp]

\centerline{\includegraphics[width=75mm,scale=1.5]{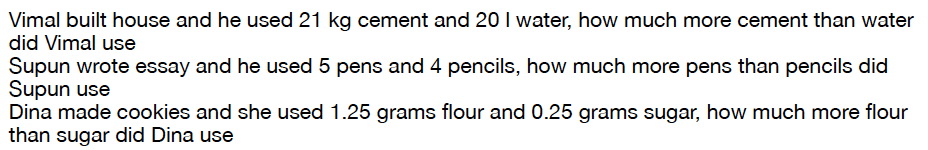}}
\caption{A sample set of English questions generated by the system (After applying constraint satisfaction with POS tags)}
\label{fig4}
\end{figure}

\section{Conclusion and Future Work}\label{Section VII}

Although many attempts have been done for Mathematics problem generation, the existing approaches are fully or semi template based approaches which restrict the creativity and novelty of  the generated questions. Our system can be identified as the first attempt to use neural language generation for the domain of automatic mathematics problem generation.

In our future work we focus on improving the accuracy of the MWPs generated by our system. In order to do that, we intend to use latest regularization and optimization techniques for LSTMs. We hope to extend our system for Tamil language MWP generation as well. Currently the system can only generate single sentenced MWPs. We hope to scale the problem generating capability of the system by introducing multiple sentence MWP generation as well.

Currently the questions generated by the system should go through a tutor to correct any minor mistakes. In order to resolve that issue, we have uses POS-based post processing, which is a hard-coded technique. As an alternative, we hope to use word embeddings as well pos tag embeddings as input features to train our neural model. 

\section*{Acknowledgement}

This research was funded by a Senate Research Committee (SRC) Grant of University of Moratuwa, Sri Lanka.

\end{document}